\pdfoutput=1
\documentclass{article}

\usepackage[preprint]{neurips_2024}

\usepackage{array,tabularx} 
\makeatletter
\providecommand\insert@pcolumn{\insert@column}
\makeatother

\usepackage[utf8]{inputenc} 
\usepackage[T1]{fontenc}    
\usepackage{hyperref}       
\usepackage{url}            
\usepackage{booktabs}       
\usepackage{amsfonts}       
\usepackage{nicefrac}       
\usepackage{microtype}      
\usepackage{xcolor}         
\usepackage{amsmath}
\usepackage{amssymb}
\usepackage{nomencl}
\usepackage{glossaries}
\usepackage{xspace}
\usepackage{multirow}
\usepackage{graphicx}
\usepackage{bbding}
\usepackage{pifont}
\usepackage{algorithm}
\usepackage{subfigure}
\usepackage{enumitem}
\usepackage{multicol}
\usepackage{caption}
\usepackage{array}
\usepackage{fp}
\usepackage{makecell}
\usepackage{flushend}
\usepackage{cases}
\usepackage{color}
\usepackage{algpseudocode}
\usepackage{listings}
\usepackage{mathrsfs}
\usepackage{longtable}
\usepackage{colortbl}
\usepackage{bm}
\usepackage{tabularx}
\usepackage{caption}
\usepackage{float}
\setlist[itemize]{noitemsep, topsep=0pt}
\definecolor{codegreen}{rgb}{0,0.3,0.6}
\definecolor{codegray}{rgb}{0.5,0.5,0.5}

\newcommand{\ignore}[1]{}

\usepackage{tcolorbox}
\definecolor{darkorange}{RGB}{255, 140, 0}
\definecolor{lightgreen}{RGB}{145, 204, 117}
\definecolor{lightyellow}{RGB}{250, 200, 88}
\definecolor{lightred}{RGB}{238, 102, 102}
\definecolor{lightblue}{RGB}{115, 192, 222} 
\newtcolorbox{promptbox}[2][Prompt]{
colback=black!5!white,
arc=5pt, 
boxrule=0.5pt,
fonttitle=\bfseries,
title=#1, 
before upper={\scriptsize}, fontupper=\fontfamily{ptm}\selectfont,
colframe=#2, 
}

\title{AD-DINOv3: Enhancing DINOv3 for Zero-Shot Anomaly Detection with Anomaly-Aware Calibration}

\author{%
  Jingyi Yuan, Jianxiong Ye, Wenkang Chen, Chenqiang Gao\thanks{ Corresponding Author}
  \vspace{3pt} \\
  School of Intelligent Systems Engineering, Sun Yat-Sen University\\
  \small{\{yuanjy36, yejx55, chenwk25\} @mail2.sysu.edu.cn} \\
  \small{$^{*}$gaochq6@mail.sysu.edu.cn}
}

\usepackage{url}
\begin{document}

\maketitle

\begin{abstract}
Zero-Shot Anomaly Detection (ZSAD) seeks to identify anomalies from arbitrary novel categories, offering a scalable and annotation-efficient solution. Traditionally, most ZSAD works have been based on the CLIP model, which performs anomaly detection by calculating the similarity between visual and text embeddings. Recently, vision foundation models such as DINOv3 have demonstrated strong transferable representation capabilities. In this work, we are the first to adapt DINOv3 for ZSAD. However, this adaptation presents two key challenges: (i) the domain bias between large-scale pretraining data and anomaly detection tasks leads to feature misalignment; and (ii) the inherent bias toward global semantics in pretrained representations often leads to subtle anomalies being misinterpreted as part of the normal foreground objects, rather than being distinguished as abnormal regions. To overcome these challenges, we introduce AD-DINOv3, a novel vision-language multimodal framework designed for ZSAD. Specifically, we formulate anomaly detection as a multimodal contrastive learning problem, where DINOv3 is employed as the visual backbone to extract patch tokens and a CLS token, and the CLIP text encoder provides embeddings for both normal and abnormal prompts. To bridge the domain gap, lightweight adapters are introduced in both modalities, enabling their representations to be recalibrated for the anomaly detection task. Beyond this baseline alignment, we further design an Anomaly-Aware Calibration Module (AACM), which explicitly guides the CLS token to attend to anomalous regions rather than generic foreground semantics, thereby enhancing discriminability. Finally, anomaly localization are achieved by measuring the similarity between adapted visual features and prompt embeddings. Extensive experiments on eight industrial and medical benchmarks demonstrate that AD-DINOv3 consistently matches or surpasses state-of-the-art methods, verifying its effectiveness and broad applicability as a general zero-shot anomaly detection framework. The code will be available at \url{https://github.com/Kaisor-Yuan/AD-DINOv3}
\end{abstract}

\section{Introduction}
Anomaly detection plays a vital role in various real-world applications, such as industrial quality inspection, medical image analysis, and security monitoring \cite{pang2021deep}. Traditional supervised anomaly detection methods typically rely on large amounts of labeled abnormal samples for each category, which is often impractical in large-scale or dynamic environments \cite{bergmann2019mvtec,akcay2018ganomaly,baur2018deep}. In contrast, Unsupervised Anomaly Detection (UAD) aims to identify anomalies in images with only normal samples for training without requiring any anomalous samples \cite{patchcore,spade,padim}. Nevertheless, it remains challenging to obtain plenty of normal images for training. 

Traditional unsupervised anomaly detection methods rely on modeling the distribution of normal samples, which makes them sensitive to domain shifts and requires retraining whenever a new product category is introduced. This limitation has motivated a growing interest in Zero Shot Anomaly Detection (ZSAD), which aims to identify anomalies in previously unseen objects without using any target-dataset samples during training. In practice, ZSAD leverages the transferability of large-scale pretrained models to generalize across categories and domains. Among them, pre-trained vision–language models (VLMs) have become particularly popular due to their strong cross-modal alignment and generalization capability. In particular, CLIP~\cite{radford2021clip}, with its powerful image–text alignment, has inspired a series of ZSAD approaches~\cite{adaclip,anomalyclip,clipad,denseclip,cocoop}. These methods typically construct anomaly maps by computing the cosine similarity between image patch features and textual prompts, thereby enabling pixel-level localization of anomalous regions.

\begin{figure}[t]
	\centering
	\includegraphics[width=1\textwidth]{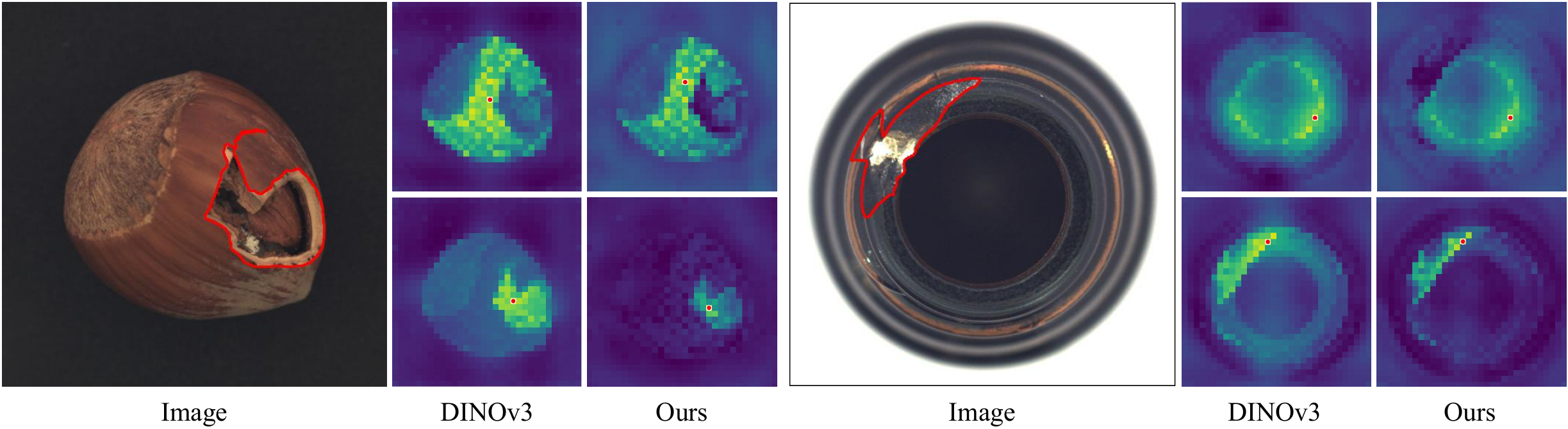}%
	\caption{Cosine similarity maps between the red-dotted patch and all other patches. The left shows the ground truth, the center corresponds to original DINOv3, and the right to our proposed AD-DINOv3. The upper and lower rows illustrate attention to normal regions and anomalous regions, respectively. Compared with original DINOv3, AD-DINOv3 reduces spurious responses in normal regions, highlights anomalous areas more prominently and coherently, and enhances the separation between normal and abnormal regions.}
	\label{fig:comparison}
\end{figure}

In recent, vision foundation models pretrained on billions of natural images have recently shown excellent cross-task transferability. To date, most ZSAD methods still rely on vision–language models such as CLIP \cite{radford2021clip}, exploiting their image–text alignment to detect anomalies. In comparison, self-supervised visual encoders like DINOv3 \cite{simeoni2025dinov3} remain largely unexplored for this task. However, the direct application of DINOv3 to ZSAD faces two key challenges: (i) the domain bias between large-scale pretraining data and anomaly detection tasks leads to feature misalignment, limiting the direct applicability of pretrained representations to anomaly detection; and (ii) the learned representations tend to emphasize global object semantics while overlooking subtle defect cues, causing anomalies to be treated as background noise rather than recognized as abnormal regions. Fig. \ref{fig:comparison} illustrates the differences between original DINOv3 and our proposed AD-DINOv3. For DINOv3, the similarity maps show two major issues. First, when attending to normal regions (upper row), the responses incorrectly extend to anomalous areas, indicating a semantic misalignment where anomalies are regarded as part of normal regions. Second, when attending to anomalous regions (lower row), the responses not only include spurious activations in normal areas but also exhibit insufficient contrast, preventing anomalies from standing out as coherent and distinctive clusters.

In this work, we propose AD-DINOv3, a novel vision–language multimodal framework and the first to adapt DINOv3 to the ZSAD task. Our approach leverages DINOv3 as the visual backbone to extract both patch tokens and CLS token enabling unified anomaly localization. To fully exploit the hierarchical representations of DINOv3, we introduce a multi-level feature adaptation strategy with lightweight adapters, preserving complementary low- and high-level cues.  Moreover, we design a novel Anomaly-Aware Calibration Module (AACM), which explicitly guides the CLS token to attend to anomalous regions rather than generic foreground semantics.  In parallel, the text branch employs the CLIP text encoder to represent both normal and abnormal prompts, which are further aligned to the target domain via adapters, performing cross modal contrastive learning with visual branch. Finally, anomaly detection is realized by comparing visual and textual representations: patch tokens against prompt embeddings for pixel-level anomaly maps. As shown in Fig. \ref{fig:comparison}, compared with the original DINOv3, our proposed AD-DINOv3 produces cleaner responses: normal regions are more compact and free from spurious activations, while anomalous regions appear more prominent and coherent. Overall, the separation between normal and abnormal regions is significantly improved.

Our main contributions include the following:
\begin{enumerate}
    \item We present AD-DINOv3, the first framework that adapts DINOv3 to zero-shot anomaly detection, bridging the gap between self-supervised visual encoders and anomaly detection tasks.
    \item We introduce a Cross Modal Contrastive Learning (CMCL) strategy with lightweight adapters to fully exploit hierarchical representations of DINOv3 for zero shot anomaly detection.
    \item We design a novel Anomaly-Aware Calibration Module (AACM), which explicitly guides the CLS token to focus on anomalous regions, alleviating its bias toward generic object semantics.
    \item Our method surpasses or matches state-of-the-art methods on eight industrial and medical benchmarks, underscoring its effectiveness as a general ZSAD framework.
\end{enumerate}
\label{sec-intro}

\section{Related Work}
\subsection{Traditional Anomaly Detection}
Traditional anomaly detection methods in computer vision typically model the distribution of normal data and treat deviations as potential anomalies. Reconstruction-based methods, such as Autoencoders and Variational Autoencoders \cite{hinton2006reducing,kingma2014vae}, attempt to reconstruct input images and use reconstruction errors to identify anomalous regions. GAN-based models further extend this idea by leveraging adversarial training to synthesize normal patterns and detect deviations \cite{akcay2018ganomaly}. Another line of work focuses on feature embedding. One-Class SVM and its deep extensions aim to learn compact decision boundaries around normal samples \cite{scholkopf2001ocsvm,ruff2018deepsvdd}. Memory-augmented approaches such as MemAE introduced external memory modules to enhance the reconstruction ability of normal data \cite{gong2019memae}. Industrial anomaly detection has also benefited from representation learning on large datasets. PatchCore \cite{roth2022patchcore} proposed a nearest-neighbor search strategy on features extracted from pretrained networks, achieving state-of-the-art performance on MVTec AD \cite{bergmann2019mvtec}. Despite their success, these methods rely heavily on abundant and clean normal data; in scenarios where data is noisy or distributions shift, their generalization ability is significantly limited.

\subsection{CLIP-based Anomaly Detection}
With the introduction of vision-language models such as CLIP, anomaly detection has been extended to a zero-shot setting. CLIP learns transferable representations by aligning image and text embeddings, enabling models to generalize to unseen categories without task-specific training.
Early attempts such as WinCLIP \cite{jeong2023winclip} explored the use of CLIP for anomaly detection by aggregating multi-scale features and aligning them with text prompts describing normal classes. Other works refined this idea by incorporating prompt learning. For instance, CoOp and its variants \cite{zhou2022coop} introduced learnable textual prompts, which were later adapted for anomaly detection to improve feature alignment. Similarly, AnomalyCLIP \cite{you2023anomalyclip} integrated multi-prompt ensembles to better capture diverse anomaly types.
More recent works introduced rectification mechanisms to address CLIP’s limitations in distinguishing subtle abnormal cues. For example, AFR-CLIP \cite{yuan2025afrclip} proposed a stateless-to-stateful anomaly feature rectification strategy, embedding defect-related information into textual prompts to improve anomaly localization. Other methods such as PromptAD \cite{han2024promptad} employed domain-specific prompt adaptation to reduce the semantic gap between textual and visual features.

While these advances highlight the potential of CLIP for anomaly detection, challenges remain in capturing fine-grained anomalies under complex backgrounds and ensuring robust performance across unseen domains. This motivates the exploration of stronger visual backbones DINOv3 and adaptive prompt-learning mechanisms tailored for anomaly detection.
\label{Related Work}

\section{Method}
\label{method}
\subsection{Problem Definition}
Given a test sample $I \in \mathbb{R}^{H \times W \times 3}$, the objective of zero-shot anomaly detection (ZSAD) is to generate a pixel-wise anomaly map $M \in [0,1]^{H \times W}$. In line with existing ZSAD frameworks, we utilize an auxiliary dataset
$D_a = \left\{ (I_1, G_1), \ldots, (I_n, G_n) \;\middle|\; G_i \in [0,1]^{H \times W} \right\}$,
which contains both normal and anomalous samples $\{I_i\}_{i=1}^n$ along with their ground-truth masks $\{G_i\}_{i=1}^n$. The model is subsequently evaluated on a disjoint test dataset $D_t = \{ I_{t_1}, \ldots, I_{t_m} \}$,
comprising unseen images from a different benchmark or domain. Importantly, to guarantee the zero-shot setting, the auxiliary and test datasets must be non-overlapping, i.e., $D_a \cap D_t = \emptyset$.

\begin{figure}[t]
	\centering
	\includegraphics[width=1\textwidth]{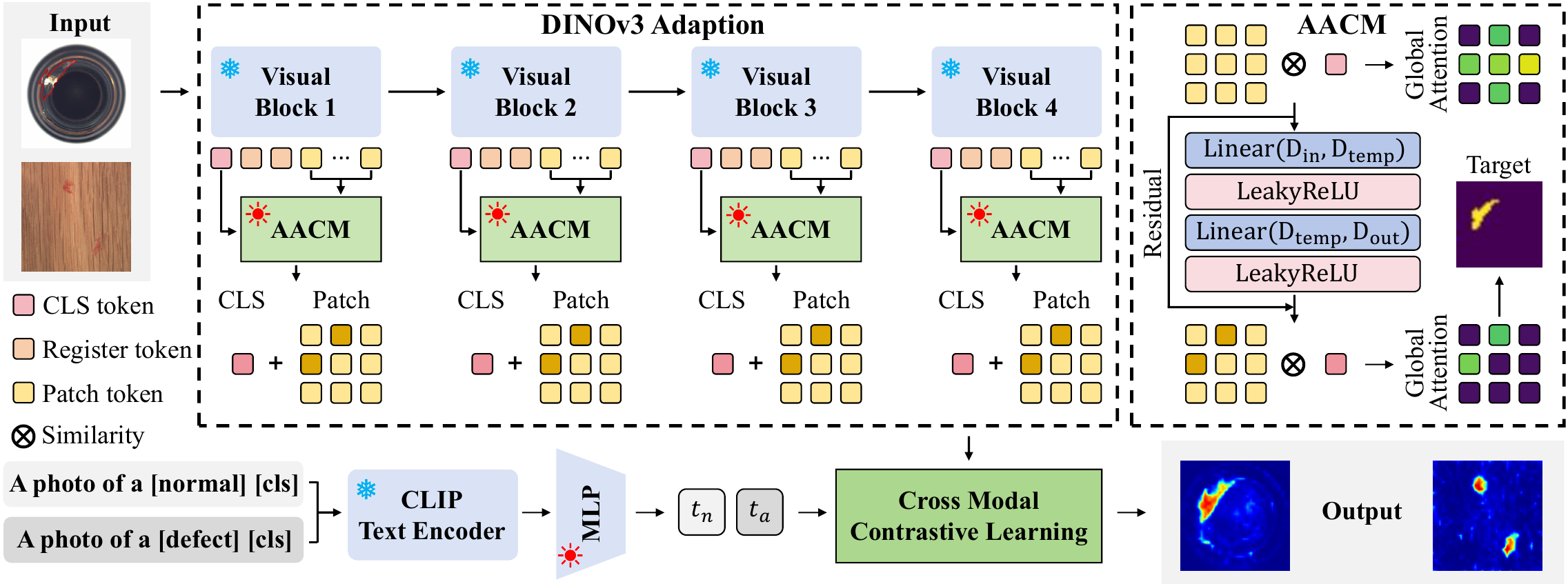} %
	\caption{The overview of our AD-DINOv3.}
	\label{fig:overview}
\end{figure}

\subsection{Overview}
We employ DINOv3 as the visual backbone of AD-DINOv3. As illustrated in Fig.~\ref{fig:overview}, the image branch extracts patch tokens and a CLS token, which are refined by lightweight adapters together with an Anomaly-Aware Calibration Module (AACM). The AACM explicitly guides the CLS token to focus on anomalous patch regions under mask supervision, redirecting its global attention from generic foreground objects, as favored in natural image pretraining, toward abnormalities. In parallel, the text branch encodes both normal and abnormal prompts (e.g., “a photo of a [state] [class]”) and is likewise refined by adapters to better align with the target domain. For anomaly localization, patch tokens are compared with the prompt embeddings to generate pixel-level anomaly maps. 

\subsection{Cross Modal Contrastive Learning (CMCL)}
Vision-language models pretrained on large-scale natural images primarily capture high-level semantics, 
such as object categories or foreground–background separation. 
While such priors are beneficial for many recognition tasks, they are not ideal for anomaly detection, 
which instead demands fine-grained discrimination between normal and abnormal regions that may differ only in subtle texture or local structure. 
Directly applying pretrained features to this task often results in weak separability: 
patches covering anomalous defects may still appear highly similar to normal counterparts, 
and textual embeddings are not explicitly aligned with anomaly semantics.

To mitigate this issue, we introduce lightweight adapters for both the visual and textual branches. 
Each adapter is implemented as a bottleneck MLP with two linear layers and LeakyReLU activations, 
which recalibrates representations toward the anomaly detection domain while keeping the powerful pretrained backbones frozen. 
We denote this transformation as $\mathrm{Ada}(\cdot)$.  

Formally, given an image, the visual encoder yields patch tokens $\{p_i\}_{i=1}^N$ and a CLS token $c$, 
while the text encoder provides embeddings for both normal and abnormal prompts $t \in \mathbb{R}^{d \times 2}$. 
The adapted representations are written as
\begin{equation}
    \{\tilde{p}_i, \tilde{c}, \tilde{t}\} = \mathrm{Ada}(\{p_i, c, t\}).
\end{equation}

We then formulate pixel-level localization as a cross modal contrastive alignment task. 
For the adapted visual patch tokens $\tilde{p}_i$, its similarity to text embeddings is computed as
\begin{equation}
    s = \frac{\tilde{p}_i^\top \tilde{t}}{\|\tilde{p}_i\|\|\tilde{t}\|}, \quad
    p = \sigma(s),
\end{equation}
where $p \in \mathbb{R}^{N\times2}$ provides normal/abnormal probabilities and $\sigma$ means softmax function over the class dimension. 
At the patch level, the abnormal probabilities are rearranged into a coarse map and bilinearly upsampled to the original image resolution, 
yielding the pixel-level anomaly map $\hat{S} \in \mathbb{R}^{H \times W}$.

\subsection{Anomaly-Aware Calibration Module (AACM)}
Although the CLS token aggregates global context by design, 
it is heavily biased toward generic foreground objects due to natural-image pretraining. 
This bias causes the model to confuse anomalies with salient object regions, 
leading to poor perception to subtle defects. 
To address this limitation, we propose the Anomaly-Aware Calibration Module (AACM), 
which introduces a mask-guided training objective that explicitly encourages the CLS token and its interaction with patch tokens to emphasize anomalous regions.

Specifically, for each patch $i$, the similarity to the adapted CLS token is given by
\begin{equation}
    s_i = \frac{\tilde{c}^\top \tilde{p}_i}{\|\tilde{c}\|\|\tilde{p}_i\|}.
\end{equation}
The similarity distribution $\sigma(\{s_i\})$ is then optimized against the anomaly mask $M$ 
using a combination of focal and Dice losses:
\begin{equation}
    \mathcal{L}_\text{AACM} = \mathcal{L}_\text{focal}(\sigma(\{s_i\}), M) 
    + \mathcal{L}_\text{dice}(\sigma(\{s_i\}), M).
\end{equation}
Here, the focal loss~\cite{focal} emphasizes hard-to-classify anomalous patches, 
while the Dice loss~\cite{dice} improves spatial consistency and alleviates the severe imbalance between normal and anomalous pixels.  

Through this objective, the CLS token is gradually guided to allocate more of its global attention to anomalous regions, 
which in turn reshapes the feature space of patch tokens associated with those regions. 
In effect, AACM enhances anomaly awareness across both global and local representations, while remaining lightweight. 

\subsection{Training and Inference}
\paragraph{Training.}
The cross-modal alignment loss is computed by comparing the cross-modal anomaly map $P$ against the ground-truth mask $M$.  
Formally,
\begin{equation}
    \mathcal{L}_\text{CM} 
    = \mathcal{L}_\text{focal}(P, M) + \mathcal{L}_\text{dice}(P, M).
\end{equation}
The complete training objective combines the cross-modal alignment loss with the AACM regularization:
\begin{equation}
    \mathcal{L} = \lambda_\text{CM}\mathcal{L}_\text{CM} + \lambda_\text{AACM}\mathcal{L}_\text{AACM},
\end{equation}
where $\lambda_\text{CM}$ and $\lambda_\text{AACM}$ are balancing weights.  

\paragraph{Testing.}
At inference time, anomaly masks are unavailable. 
Detection therefore relies solely on cross-modal similarity: 
patch–text similarities yield pixel-level anomaly maps.
Since AACM is used only during training, our framework maintains the same efficiency as the baseline model at test time.

\section{Experiments}
\label{sec:experiments}
\subsection{Experiments Setup}
\paragraph{Datasets.} To comprehensively assess the effectiveness of AD-DINOv3, we perform experiments in both industrial and medical scenarios. For the industrial setting, we adopt the MVTec AD~\cite{mvtec}, VisA~\cite{visa}, BTAD~\cite{btad}, and MPDD~\cite{mpdd} benchmarks, while in the medical setting we evaluate on ISIC~\cite{isic}, ColonDB~\cite{colon}, ClinicDB~\cite{clinic}, and TN3K~\cite{tn3k}. Since the object categories in VisA are distinct from those in the other datasets, we follow a cross-dataset evaluation strategy: model trained on VisA is tested on the remaining datasets, and conversely, we utilize the model trained on MVTec AD for VisA testing.

\paragraph{Evaluating Metrics.} Consistent with prior works in zero-shot anomaly detection, we adopt several evaluation metrics for anomaly localization. Specifically, pixel-level anomaly localization is assessed by the Area Under the Receiver Operating Characteristic Curve (AUROC), which measures the discriminative capability and the balance between precision and recall. In addition, to ensure fair comparison with existing ZSAD approaches, we include the maximum F1 score (F1). Finally, to provide a holistic assessment, we also summarize the mean performance across all domains.

\paragraph{Implementation Details.} In our experiments, we employ the pretrained DINOv3 with a ViT-L/16 backbone released by Meta AI as the default image encoder, while the text encoder from pretrained \textbf{CLIP} (OpenAI) is used to generate textual embeddings. All input images are uniformly resized to $512 \times 512$ for both training and inference. The DINOv3 backbone contains 24 transformer layers, which we partition into four stages, extracting patch embeddings from the 6th, 12th, 18th, and 24th layers, respectively. 
Model training is performed for 10 epochs with a batch size of 64, optimized using Adam. The learning rate is initialized at $1 \times 10^{-4}$. All experiments are conducted on a single NVIDIA RTX A6000 GPU (48 GB).

\paragraph{Comparison Methods.} For a comprehensive comparison, we benchmark AD-DINOv3 against representative approaches from two distinct paradigms: those that operate in a training-free manner and those that rely on auxiliary datasets. Within the first paradigm, we consider WinCLIP~\cite{winclip}, a pioneering CLIP-based framework that dispenses with any additional training data. Within the second paradigm, we examine APRIL-GAN~\cite{aprilgan}, AdaCLIP~\cite{adaclip}, and AnomalyCLIP~\cite{anomalyclip}, which are trained on external auxiliary data and subsequently evaluated on unseen object categories.

\begin{table}[b]
  \vspace{-1.em} 
  \centering
  \caption{Pixel-level performance comparisons of the ZSAD methods on the industrial and medical datasets. The best performance is in \textbf{bold} and the second-best is \underline{underlined}.}
  \setlength{\tabcolsep}{1mm}  
  \small  
  \begin{tabular}{lccccccccccc}
    \toprule[1.5pt]
    \multirow{2}{*}{Domain} & \multirow{2}{*}{Dataset} & \multicolumn{2}{c}{WinCLIP} & \multicolumn{2}{c}{APRILGAN} & \multicolumn{2}{c}{AnomalyCLIP} & \multicolumn{2}{c}{AdaCLIP} & \multicolumn{2}{c}{\textbf{Ours}} \\
    \cmidrule(lr){3-4}\cmidrule(lr){5-6}\cmidrule(lr){7-8}\cmidrule(lr){9-10}\cmidrule(l){11-12}
    & & AUROC & F1  & AUROC & F1  & AUROC & F1  & AUROC & F1  & AUROC & F1 \\
    \midrule
    \multirow{5}{*}{Industrial} 
    & VisA    & ( 79.6 ~, &14.8 )  &( 95.2 ~,  &32.3  ) &( 95.4 ~,  & 28.3 ) &( \underline{95.5} ~,  & \underline{37.7} )  &( \textbf{95.6} ~, & \textbf{37.7} )  \\
    
    & BTAD    & ( 72.6 ~, &18.5 )  &( 89.5 ~,  &38.4 ) &( \textbf{94.0} ~,  & 49.7 ) &( 92.1 ~,  &\underline{51.7} )  &( \underline{93.5} ~, & \textbf{51.7} )  \\

    & MPDD    & ( 71.2 ~, &15.4 )  &( 95.1 ~,  &30.6 ) &( \textbf{96.4} ~,  & 34.2 ) &( 96.1 ~,  &\underline{34.9} )  &( \underline{96.2} ~, & \textbf{38.8} )  \\
    
    & MVTec AD & ( 85.1 ~, & 31.6 )  &( 83.7 ~,  & 39.8 ) &( \underline{90.9} ~,  & 39.1 ) &( 88.7 ~,  & \underline{43.4} )  &( \textbf{91.6} ~, & \textbf{50.1} ) \\
    \cmidrule(l){2-12}
    
    &\textit{Average} & ( 77.1 ~, &20.1 )  &( 90.9 ~,  &35.3 ) &( \underline{94.2} ~,  & 37.5 ) &( 93.1 ~,  &\underline{41.9} )  &( \textbf{94.2} ~, & \textbf{44.6} )  \\
    
    \midrule
    \multirow{5}{*}{Medical}    
    & ISIC    & ( 83.3 ~, & 48.5 )  &( 85.8 ~,  & 67.3 ) &( \underline{89.7} ~,  & 70.6 ) &( \textbf{90.3}~,  & \textbf{72.6} )  &( 89.0 ~, & \underline{72.1} )  \\
    
    & ColonDB & ( 70.3 ~, & 19.6 )  &( 78.4 ~,  & 33.2 ) &( \underline{81.9} ~,  & \underline{37.3} ) &( \textbf{82.6} ~,  & 36.1 )  &( 80.8 ~, & \textbf{38.4} )  \\
                                
    & ClinicDB & ( 51.2 ~, & 24.4 )  &( \underline{83.2} ~,  & \underline{42.3} ) &( 82.9 ~,  & 42.1 ) &( 82.8 ~,  & 40.9 )  &( \textbf{90.4} ~, & \textbf{54.3} )  \\

    & TN3K & ( 70.7 ~, & 30.0 )  &( 74.4 ~,  & 39.7 ) &( \textbf{81.5} ~,  & \textbf{47.9} ) &( 76.8 ~,  & 40.7 )  &( \underline{77.8} ~, & \underline{41.1} )  \\

    \cmidrule(l){2-12}
    
    &\textit{Average} & ( 68.9 ~, & 30.6 )  &( 80.5 ~,  & 45.6 ) &( \underline{84.0} ~,  & \underline{49.5} ) &( 83.1 ~,  & 47.6 )  &( \textbf{84.5} ~, & \textbf{51.5 })  \\
    \bottomrule[1.5pt]
  \end{tabular}
  \label{tab:pixel-level}
\end{table}

\begin{figure}[htbp]
	\centering
	\includegraphics[width=0.9\textwidth]{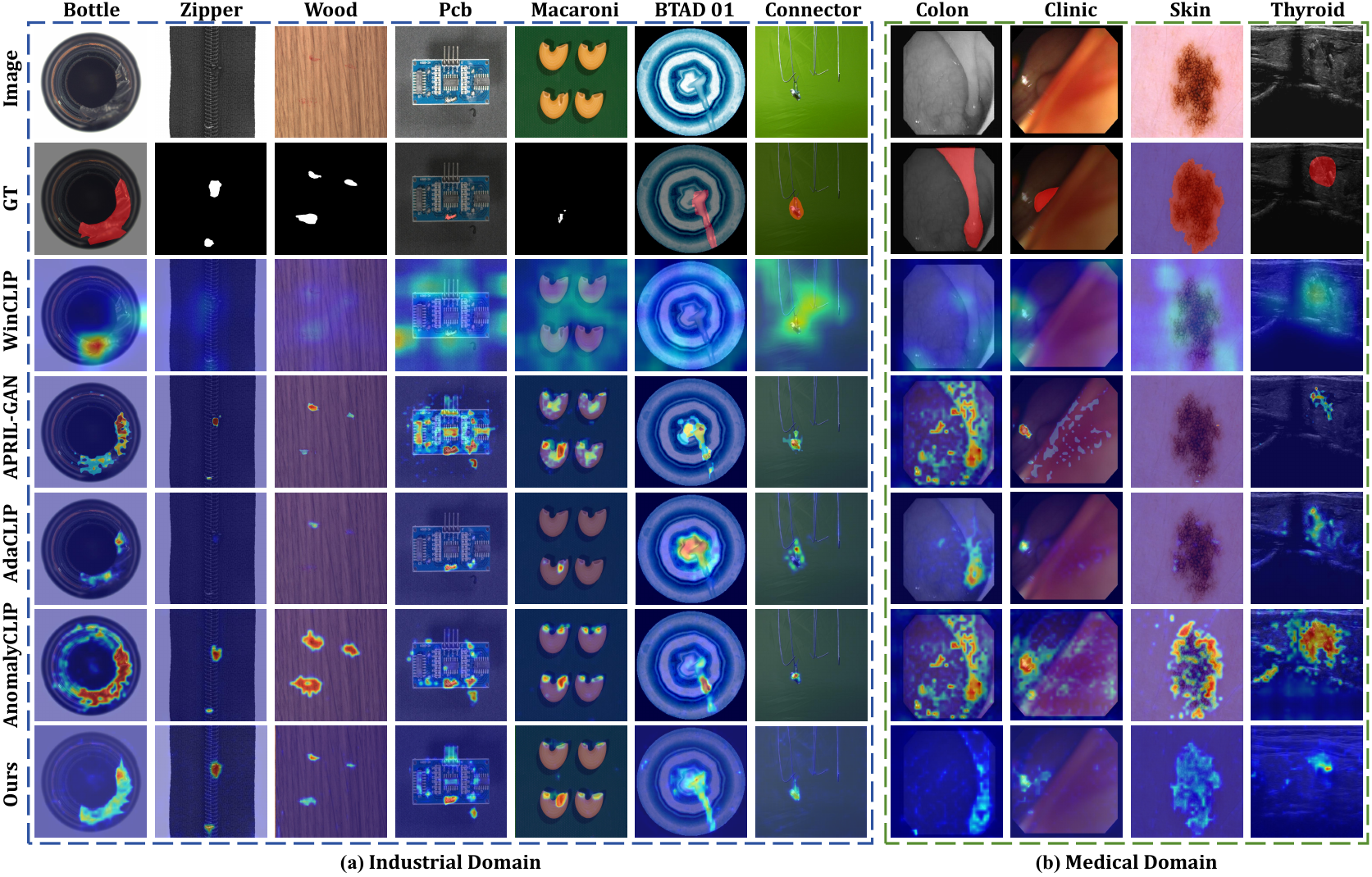}%
	\caption{Visualization of ZSAD methods. (a) and (b) present the localization results across industrial and medical domains, respectively.}
	\label{fig:visual}
    \vspace{-0.5em} 
\end{figure}

\subsection{Comparison with State-of-the-Art Methods}

\textbf{Comparison on industrial image benchmarks.}

As shown in Tab. \ref{tab:pixel-level}, AD-DINOv3 achieves state-of-the-art performance on industrial benchmarks. On VisA, our method reaches an AUROC of 95.6\% and an F1 score of 37.7\%, surpassing previous methods including WinCLIP (79.6\% / 14.8\%) and AnomalyCLIP (95.4\% / 28.3\%). Similarly, on BTAD and MPDD, our approach yields AUROC scores of 93.5\% and 96.2\%, respectively, consistently outperforming AdaCLIP and APRILGAN. On the widely-used MVTec AD, AD-DINOv3 achieves 91.6\% AUROC and 50.1\% F1, setting a new benchmark.

When averaged across all industrial datasets, AD-DINOv3 attains an AUROC of 94.2\% and an F1 score of 44.6\%, clearly exceeding all competing methods. This demonstrates the advantage of incorporating DINOv3’s discriminative representations, which enhance the separability between normal and anomalous features.

Qualitative comparisons in Fig. \ref{fig:visual}(a) further confirm these findings. Competing approaches, particularly WinCLIP and APRILGAN, often produce blurry or noisy heatmaps, failing to accurately delineate defect regions. In contrast, AD-DINOv3 generates sharp and precise anomaly maps, effectively highlighting defective components such as scratches on metal surfaces and structural defects in PCB boards, while suppressing irrelevant background noise.

\textbf{Comparison on medical image benchmarks.}

We further evaluate AD-DINOv3 on medical datasets to assess its ability to handle fine-grained and visually subtle anomalies. As summarized in Tab.~\ref{tab:pixel-level}, our method achieves consistent improvements over prior works. On ClinicDB, AD-DINOv3 delivers an AUROC of 90.4\% and an F1 score of 54.3\%, significantly outperforming AnomalyCLIP (82.9\% / 42.1\%) and AdaCLIP (82.8\% / 40.9\%). On ISIC, our method achieves an AUROC of 89.0\% with an F1 score of 72.1\%, competitive with AdaCLIP while maintaining balanced localization accuracy. Similar gains are observed on ColonDB and TN3K, where our approach consistently improves upon existing CLIP-based baselines.

On average, across the four medical datasets, AD-DINOv3 achieves an AUROC of 84.5\% and an F1 score of 51.5\%, establishing new state-of-the-art performance. This highlights the robustness of our framework in capturing anomalies in medical images, which are typically small in size and highly ambiguous.

The qualitative results in Fig. \ref{fig:visual}(b) further illustrate this advantage. Competing approaches often misclassify normal tissues as abnormal or fail to localize subtle lesions, especially in dermoscopic and endoscopic images. Conversely, AD-DINOv3 generates clearer and more discriminative heatmaps, successfully identifying lesions in skin and colon images and localizing irregular patterns in thyroid scans. These results underscore the potential of our method for real-world medical applications where high precision and sensitivity are critical.

\subsection{Ablation Study}
As shown in Tab.~\ref{tab:ablation} we dismantle AD-DINOv3 step-by-step on MVTec AD to verify the contribution of each component. For the baseline, We simply $\ell$2-normalize the raw DINOv3 patch tokens, average over channel dimension, and upsample to the image size as the anomaly map. This yields 76.20\% AUROC and 20.49\% F1, confirming that generic self-supervised features alone cannot reliably separate normal and anomalous regions.
\paragraph{Ablation study on the Cross-Modal Contrastive Learning (CMCL).} 
As shown in row 2 of Tab.~\ref{tab:ablation}, introducing CMCL, which explicitly aligns visual and textual features, leads to a large performance gain: AUROC rises to 90.98\% (+14.78\%) and F1 to 47.00\% (+26.51\%). 
This substantial improvement indicates that cross-modal alignment effectively narrows the gap between image features and semantic prompts. 
By enforcing consistency across modalities, CMCL calibrates the embedding space so that normal and abnormal regions become more separable, thereby refining pixel-level decision boundaries and enhancing overall discriminability. 
These results confirm that leveraging semantic cues from the text branch is crucial for guiding the visual representations toward anomaly-aware feature learning.

\paragraph{Ablation study on the Anomaly-Aware Calibration Module (AACM).} 
As shown in row 3 of Tab.~\ref{tab:ablation}, adding the proposed AACM further increases AUROC to 91.60\% and F1 to 50.13\%. 
This gain demonstrates that explicitly guiding the global CLS token to attend to defective regions enhances.
By redirecting attention from generic foreground semantics to anomalies, AACM calibrates the representation space and makes the model focus on subtle yet critical cues, thereby yielding more accurate anomaly maps and more reliable image-level decisions.

\begin{table}[t]
\vspace{-2.em} 

\centering
\small
\caption{Ablation Results of AD-DINOv3's Components on MVTec AD.}
\label{tab:ablation}
\resizebox{0.65\linewidth}{!}{
\begin{tabular}{ccccccc}
\toprule
Num.& CMCL & AACM & AUROC & F1\\
\midrule
  1.& \ding{55} & \ding{55} & 76.20 & 20.49 \\
  2.& \ding{52} & \ding{55} & 90.98 (+14.78) & 47.00 (+26.51) \\
  3.& \ding{52} & \ding{52} & 91.60 (+15.40) & 50.13 (+29.64) \\

\bottomrule
\end{tabular}}
\end{table}

\begin{table}[t]
\centering
\small
\caption{Ablation Results of multi-layer features on MVTec AD.}
\label{tab:multi-layer}
\resizebox{0.7\linewidth}{!}{
\begin{tabular}{ccccccc}
\toprule
Num.& Last Layer & Multi Layer & AUROC & F1\\
\midrule
  1.& \ding{52} &  & 90.38 & 48.24 \\
  2.&  & \ding{52} & 91.60 (+1.22) & 50.13 (+1.89) \\

\bottomrule
\end{tabular}}
\end{table}

\textbf{Ablation study on multi-level features.}
To investigate the effect of utilizing features from different layers of the visual backbone, we conduct an ablation study comparing single-layer and multi-layer settings. In the single-layer variant, only the last-layer output is used to compute similarity with text embeddings for anomaly detection. In contrast, the multi-layer variant aggregates features from the 6th, 12th, 18th, and 24th layers, and the anomaly maps from these four stages are averaged to obtain the final prediction.

As shown in Tab.~\ref{tab:multi-layer}, incorporating multi-level features leads to consistent improvements across all evaluation metrics. Compared with the single-layer baseline, the multi-layer strategy achieves an increase of +1.22\% in AUROC and +1.89\% in F1 score, demonstrating that intermediate features contain complementary information that helps capture both low-level appearance cues and high-level semantic context. This indicates that leveraging hierarchical representations enhances the model’s ability to distinguish anomalies from normal regions.

\section{Conclusion}
In this work, we introduced AD-DINOv3, the first framework that adapts DINOv3 to the ZSAD task. Our approach integrates lightweight adapters and a novel Anomaly-Aware Calibration Module (AACM) to refine patch and CLS tokens, explicitly guiding the model to focus on anomalous regions rather than generic foreground semantics. Across 8 industrial and medical benchmarks, AD-DINOv3 matches or surpasses state-of-the-art approaches, underscoring its robustness and broad applicability to ZSAD. We further observe that DINOv3 representations fragment normal regions into multiple clusters due to intra-class variations, while anomalies fail to form a distinctive cluster. Overall, our study shows that DINOv3 yields more discriminative visual features than CLIP, offering a powerful and generalizable solution to ZSAD.


\bibliographystyle{unsrt}
\bibliography{ref.bib}

\end{document}